\documentclass{article}

\PassOptionsToPackage{numbers, compress}{natbib}
\usepackage{aimc2022}

\usepackage[utf8]{inputenc} 
\usepackage[T1]{fontenc}    
\usepackage{hyperref}       
\usepackage{url}            
\usepackage{booktabs}       
\usepackage{amsfonts}       
\usepackage{nicefrac}       
\usepackage{microtype}      
\usepackage{graphicx}       

\usepackage{graphicx}
\usepackage{tikz}
\usepackage{comment}
\usepackage{amsmath,amssymb,bm} 
\usepackage{bbding}
\usepackage{color}
\usepackage{multirow}
\usepackage{wrapfig}

\usepackage[accsupp]{axessibility}  

\title{Test-time Batch Normalization}

\author{Tao Yang$^1$\thanks{Work done during internships at Microsoft Research Asia.}
, Shenglong Zhou$^{3}$
, Yuwang Wang$^2$\thanks{Corresponding author}  
, Yan Lu$^2$ 
, Nanning Zheng$^1$\\
\texttt{yt14212@stu.xjtu.edu.cn},\\
\texttt{slzhou96@mail.ustc.edu.cn}, \\ \texttt{\{yuwwan,yanlu\}@microsoft.com},\\ \texttt{nnzheng@mail.xjtu.edu.cn}\\
$^1$Xi'an Jiaotong University, $^2$Microsoft Research Asia,\\
$^3$University of Science and Technology of China\\
}

\begin{document}

\maketitle

\begin{abstract}
Deep neural networks often suffer the data distribution shift between training and testing, and the batch statistics are observed to reflect the shift. In this paper, targeting of alleviating distribution shift in test time, we revisit the batch normalization (BN) in the training process and reveals two key insights benefiting test-time optimization: $(i)$ preserving the same gradient backpropagation form as training, and $(ii)$ using dataset-level statistics for robust optimization and inference. Based on the two insights, we propose a novel test-time BN layer design, GpreBN, which is optimized during testing by minimizing Entropy loss. We verify the effectiveness of our method on two typical settings with distribution shift, i.e., domain generalization and robustness tasks. Our GpreBN significantly improves the test-time performance and achieves the state of the art results.
\end{abstract}

\section{Introduction}

In recent years, deep learning has achieved remarkable success in various tasks when the training and testing data follow the independent and identical distribution (i.i.d.) assumption~\cite{ren2015faster,he2016deep,minaee2021image}. However, this assumption may not hold in real-world applications, and the models are often sensitive to the distribution shift between the training (source) and testing (target) data~\cite{recht2019imagenet,gulrajani2020search}. 
Some approaches have been explored in the training and testing stages to address the challenge. In the training stage, many efforts have been spent on learning domain-invariant or robust representation that can be generalized on target data~\cite{krueger2021out,lu2020stochastic,li2021learning,muandet2013domain,cha2021swad,zhou2021domain}. However, the testing distribution is not always accessible during training. Tackling the distribution shift at test time is more applicable. 
Recently, several test-time approaches have been proposed to adapt the models to alleviate the domain shift with testing data available~\cite{sun2020test,wang2020tent}.

\begin{figure}
    \centering
    \includegraphics[width=\linewidth]{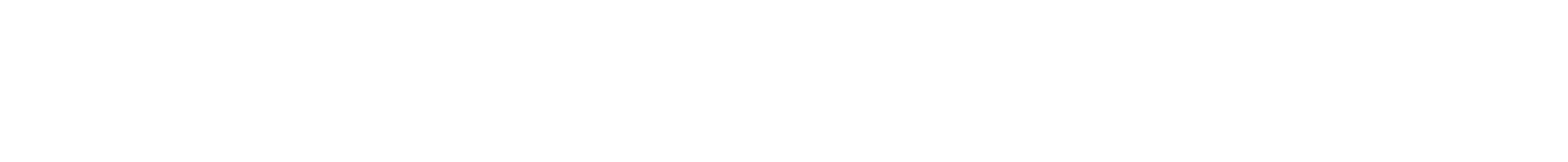}
    \label{fig:overview}
\end{figure}

We are particularly interested in the test-time solutions for the potential of well addressing the domain shift for given trained models. The BN layer~\cite{ioffe2015batch} is found to be highly related to the model performance in the presence of data distribution shift~\cite{pan2018two,schneider2020improving,nado2020evaluating,wang2020tent,you2021test,seo2020learning}. There are two sets of parameters inside the standard BN layer, the statistics ($\mu,\sigma$) used to normalize the feature and the affine parameters ($\beta,\gamma$) to modulate the feature. Some works~\cite{benz2021revisiting,schneider2020improving,wang2020tent} simply substitute the source batch statistics with the statistics of the current batch on the target domain. Some other works mix the statistics of source and target~\cite{you2021test,nado2020evaluating}. 
Tent~\cite{wang2020tent} and its following works~\cite{you2021test,khurana2021sita,hu2021mixnorm,zhang2021test} optimize the affine parameters via minimizing Entropy loss in test time. 

All these methods focus on the test-time adaptation of BN. Before that, we would like first to revisit BN in the training stage. We follow BN implemented in Pytroch~\cite{paszke2017automatic} as a standard tool.
Given a batch of input of BN layer $\{x_i\in \mathbb{R}^d\}_{i=1}^B$, where $d$ is the feature dimension, and $B$ is batch size. In the training phase, the outputs $\{y_i\in \mathbb{R}^d\}_{i=1}^B$ of the BN layer are calculated as
\begin{equation}
    {y_i} = \frac{x_i - \mu}{\sigma} \gamma + \beta;\; \mu = \mathbb{E}[x_i];\; \sigma^2 = \mathbb{E}[(x_i - \mu)^2].
    \label{eq1}
\end{equation}
Meanwhile, a running mean vector $\mu_r$ and a running variance vector $\sigma_r$ are estimated using moving average:
\begin{equation}
    \mu_r = \lambda \mu + (1-\lambda) \mu_r;\; \sigma_r^2 = \lambda \sigma^2 + (1-\lambda) \sigma_r^2,
    \label{eq2}
\end{equation}
where $\lambda = 1/n$, and $n$ is the number of batches tracked\footnote{Here we take the cumulative moving average as an example, and the same goes for another common alternative, exponential moving average ($\lambda \in (0,1)$ is fixed).}. $\mu_r, \sigma_r$ are used to normalize the input in the testing phrase:
\begin{equation}
    {y_i} = \frac{x_i - \mu_r}{\sigma_r} \gamma + \beta.
    \label{eq3}
\end{equation}


Revisiting the training stage, we get two important insights. The first one is on the gradient backpropagation. Since the BN layer uses $\mu,\sigma$ of the current batch, the backpropagated gradient of one input can be impacted by another one via the gradient of $\mu$ and $\sigma$. 
We empirically show that, in test-time adaptation,  preserving this cross instance gradient backpropagation is crucial. 
To the best of our knowledge, we are the first to analyze test-time BN from the perspective of gradient backpropagation.
Our second insight is that the normalization statistics should be dataset-level. This is inspired by the fact that we use the running averaged statistics $\mu_r$ and $\sigma_r$ for testing in the i.i.d. setting. The dataset-level statistics are more stable and robust for inference and optimization. 


However, the gradient backpropagation form (computational graph) and the selection of normalization statistics are often coupled in the BN layer. The choice of statistics can impact the form of gradient backpropagation. For example, if we use the source statistics $\mu_r, \sigma_r$, due to $\mu_r, \sigma_r$ being fixed and irrelevant with target data, the gradient backpropagation is not cross the current batch.
To preserve the backpropagation form and allow for flexible selection of statistics, we propose a novel form of BN layer, denoted as GpreBN. With GpreBN, we can choose the moving average statistics or other alternatives of statistics flexibly, while still preserving the backpropagation form. 

To verify the effectiveness of our proposed GpreBN, inspired by~\cite{wang2020tent}, we also choose Entropy as the minimizing loss function to optimize the affine parameter of the BN layer. Most of the previous methods are limited to only domain generalization or robustness settings. For example, we empirically show that Tent works well on robustness but causes a significant drop on domain generalization task. We verify the effectiveness of our method on both domain generalization and robustness settings. In both tasks, we achieve state-of-the-art results, demonstrating that our method provides a general solution that can be applied well to address the data distribution shift. 

Our main contributions can be summarized as follows:
\begin{itemize}
    \item We present two insights for test-time BN: $(i)$ preserving backpropagation form of training stage, $(ii)$ using dataset-level statistics.
    \item We propose a novel test-time GpreBN to effectively address distribution shift based on the previous two insights. 
    \item We conduct experiments on domain generalization and robustness settings to verify the effectiveness of our method.
\end{itemize}

\section{Related works}
In this section, we review the related works on tackling data distribution shift between training and testing. For domain adaptation and generalization, the shift is the domain gap. For the robustness task, the distribution shift is also known as covariate shift, which is simulated by perturbations of the testing data. We also introduce some other related test-time adaptation works. 

\paragraph{\textbf{Batch Normalization}}
Batch normalization (BN) is widely used in deep networks, e.g., Resnet~\cite{he2016deep}, Wide Resnet~\cite{zagoruyko2016wide}, which is originally proposed to address the internal covariate shift between different training batches for fast and stable converge~\cite{ioffe2015batch}. There are a lot of works on BN. Here we only focus on the works related to the out-of-distribution issue.
Some works find that only adapting the batch statistics is effective for domain adaptation or robustness.
For example, Li et al.~\cite{li2016revisiting} use target domain batch statistics for domain adaptation. The recent works ~\cite{nado2020evaluating,schneider2020improving} improve the robustness to common corruptions by re-calculating the batch statistic on testing data or mixing with source domain statistics. Some works~\cite{pan2018two,seo2020learning} combine BN with other types of normalization, e.g., instance normalization, to provide more statistics information to address distribution shift. The previous works do not adapt the model or only adapt the model with the predefined operation, while our work focuses on BN layer design targeting for test-time optimization.

\paragraph{\textbf{Domain adaptation, Domain generalization, and Robustness}} In order to handle the domain gap, the research community has spent significant effort on domain adaptation~\cite{krueger2021out,lu2020stochastic,li2021learning} and domain generalization~\cite{muandet2013domain,cha2021swad,zhou2021domain}. Domain adaptation requires the access of target data during training. Domain generalization aims to generalize to unseen target domains without access to the target domain. Most of these methods are training stage approaches, and the trained model is not adapted during testing. They target training the feature extractor to learn domain-invariant representation~\cite{muandet2013domain,seo2020learning,zhao2020domain,li2018domain} or using data augmentation~\cite{shankar2018generalizing,volpi2018generalizing,zhou2021domain} to alleviate the domain gap.  
For robustness, researchers investigate developing methods to handle different perturbations~\cite{dodge2017study,hendrycks2021many,hendrycks2019augmix}. ~\cite{lakshminarayanan2017simple,ovadia2019can} explore the calibration of the model or classifier to make the model robust to perturbations. In addition, data augmentation-based methods~\cite{gowal2021improving,hendrycks2019augmix} are popular in the literature. However, the distribution shift may not be completely removed due to the inaccessibility of target data. Different from the previous training stage methods, we target a domain generalization setting to verify the effectiveness of handling distribution shift only in test-time.

\paragraph{\textbf{Test-time Adaptation}} Some approaches propose to adapt the trained model in test-time. This branch is originated from the works of re-calculating the batch statistics~\cite{li2016revisiting}, as reviewed in the batch normalization section. 
TTT~\cite{sun2020test} adapts the entire model, which is trained with both a supervised and a self-supervised loss, using the self-supervised loss on the testing set. Tent~\cite{wang2020tent} demonstrates that minimizing the Entropy loss via optimizing the affine parameters of BN is effective for robustness and source-free domain adaptation tasks. The following works~\cite{khurana2021sita,hu2021mixnorm,zhang2021test} study different test-time settings, e.g., single test sample. Besides these methods on robustness, following Tent, You et al.~\cite{you2021test} replace the target statistics used in Tent by mixing source and target statistics. T3A~\cite{iwasawa2021test} adapts the classifier in test-time by adopting centroid-based modification for domain generalization. Based on the previous works, we also adapt the BN layer via minimizing Entropy loss in test-time. Different from the previous methods, We reveal two novel insights: $(i)$ preserving the gradient backpropagation form the same as in the training stage, and $(ii)$ using dataset-level statistics, which is implemented with moving average following the training stage. In the domain adaptation setting,  AdaBN~\cite{li2016revisiting} proposes to use the statistics calculated on the available target domain data. However, it does not further update the statistics during testing. Besides, our proposed BN layer decouples the form of gradient backpropagation and the statistics used for normalization.

\section{Test-time Batch Normalization}
In this section, we would like first to revisit Batch Normalization in the training stage and demonstrate the necessity of preserving cross instance gradient propagation in Section~\ref{sec:revisit}. We then present our proposed GpreBN in Section~\ref{sec:GpreBN}. Several batch statistics options are presented in Section~\ref{sec:rBN}.

\subsection{Revisiting Batch Normalization}
\label{sec:revisit}

Batch normalization (BN) has been extensively used in deep learning to achieve faster converge and better performance. As presented in Equation~\ref{eq1},~\ref{eq2},~\ref{eq3} in the introduction section, the standard BN layer relies on the batch-level statistics $\mu,\sigma$ to normalize the feature during training. From the forward propagation perspective, BN can reduce internal variance between different batches~\cite{ioffe2015batch}. In this paper, we would like to provide an interesting view from the backpropagation. The input and output of the BN layer are $\{x_i\in \mathbb{R}^d\}_{i=1}^B$ and $\{
y_i\in \mathbb{R}^d\}_{i=1}^B$, respectively. Assuming the loss function is $\mathcal{L}$, the backpropagated gradient of $\mathcal{L}$ on ${y_i}$ is $\partial \mathcal{L}/\partial {y_i}$. According to Equation~\ref{eq1}, the gradient of $\mathcal{L}$ on the input ${x_i}$ is
\begin{equation}
    \frac{\partial \mathcal{L}}{\partial x_i} = \frac{\partial \mathcal{L}}{\partial {y_i}}\frac{\partial}{\partial x_i}\left(\frac{x_i-\mu}{\sigma}\right)\gamma.
    \label{eq4}
\end{equation}
Since $\mu, \sigma$ is calculated from the batch samples, the gradient backpropagation is cross-instance, i.e., the gradient of $x_i$ is impacted by $y_j$, even $(j\neq i)$. 

Turing to test-time adaptation, the network parameters are updated by backpropagation. One may use different statistics. For example, the running average statistics is used for normalization, i.e., $\mu=\mu_r, \sigma=\sigma_r$. Since they are calculated on the source data and irrelevant with the inputs in test-time, the gradients of $x_i$ is
\begin{equation}
    \frac{\partial \mathcal{L}}{\partial x_i} = \frac{\partial \mathcal{L}}{\partial {y_i}}\frac{\gamma}{\sigma_r}.
    \label{eq:partial_r}
\end{equation}
Compared to Equation~\ref{eq4}, there is no cross-instance backpropagation. 

\begin{table*}[t]
    \caption{Comparison on performance of Tent on domain generalization (PACS) and robustness (CIFAR-10-C) with different statistics. The results of using source statistics (source) are presented in left column, and target statistics (target) are presented in right column. Tent(de) represent Tent with stopping gradient of $\mu,\sigma$.}
    \centering
    \begin{tabular}{lcc|ccc}
    \hline
    \textbf{Dataset}   &source &+Tent&target&+Tent&+Tent(de)\\
    \hline
    PACS (accuracy$\uparrow$)        &84.7 &84.6 & 85.6 &90.2 &87.6  \\
    CIFAR-10-C (error$\downarrow$)       & 11.2  & 17.7
    & 10.9 &9.2&10.3\\
    \hline
    \end{tabular}
    \label{Intro_tab}
    \vspace{-1.5em}
\end{table*}

We empirically study the impact of this discrepancy on backpropagation. We choose Tent~\cite{wang2020tent} as its effectiveness in increasing the robustness in test-time adaptation. Specifically, Tent~\cite{wang2020tent} optimizes the affine parameters $\beta, \gamma$ by minimizing Entropy loss. Following the default setting, Tent uses the statistics of the current target testing batch, i.e., $\mu = \mathbb{E}[x_i]$ and $\sigma^2 = \mathbb{E}[(x_i - \mu)^2]$, $x_i$ is from testing data. Consequently, there is cross-instance backpropagation. To simulate the propagation in Equation~\ref{eq:partial_r}, we detach the gradient backpropagation on $\mu,\sigma$ and keep others the same. The result is shown in Table~\ref{Intro_tab}. Surprisingly, the gain of Tent drops a lot when there is no cross instance backpropagation (target+Tent(de)) compared to the raw implementation (target+Tent) in both domain generalization (gain drops from 4.6\% to 2.6\%) and robustness (gain drops from 1.7\% to 0.6\%) tasks. 
If we replace the target statistics with source statistics ($\mu_r,\sigma_r$) by optimizing the affine parameters in BN, there is almost no gain or even a significant drop. 
In a nutshell, we reveal that preserving the cross-instance backpropagation as training is crucial for test-time adaptation of the BN layer. 
\subsection{Gradient Presevering Batch Normalization}
\label{sec:GpreBN}
However, in test-time, statistics of the current batch used in Tent may not be the optimal option to normalize the input. 
Based on the observation above, in order to use different normalizing statistics while preserving the cross instance gradient backpropagation, we propose a novel BN form, GpreBN, as
\begin{equation}
    {y_i} = \frac{\frac{x_i-\mu_c}{\sigma_c}\bar{\sigma_c} + \bar{\mu_c} - \mu}{\sigma}\gamma + \beta,
    \label{equ:gprebn}
\end{equation}
where $\mu_c=\mathbb{E}[x_i]$ and $\sigma_c=\mathbb{E}[(x_i - \mu)^2]$ are the statistics of the current batch, and ${\bar{\mu}}_c$, ${\bar{\sigma}}_c$ denote stopping gradient backpropagation. We leave $\mu$ and $\sigma$ to denote the arbitrary non-learnable parameters to normalize the feature. The backpropagation of our GpreBN is 
\begin{equation}
  \frac{\partial \mathcal{L}}{\partial x_i} = \frac{\partial \mathcal{L}}{\partial {y_i}}\frac{\partial}{\partial x_i}\left(\frac{x_i-\mu_c}{\sigma_c}\right)\frac{\bar{\sigma_c}}{\sigma}\gamma.
    \label{equ:gprebn_bp}
\end{equation}
Note that the cross-instance gradient backpropagation is preserved, and there is only one multiplier difference between this gradient and Equation~\ref{eq4}. In summary, GpreBN decouples the backpropagation form and the selection of normalization statistics: the backpropagation is implemented via $\mu_c,\sigma_c$, normalization statistics is implemented via $\mu,\sigma$. Note that the BN form of Tent is a special case of GpreBN when we choose $\mu=\bar{\mu}_c,\sigma=\bar{\sigma}_c$. 


\subsection{Options of Normalization Statistics}
\label{sec:rBN}
In this section, we discuss the options of statistics ($\mu,\sigma$ in Equation~\ref{equ:gprebn}) used to normalize the input in GpreBN. One may observe that the value of $\mu, \sigma$ in GpreBN (Equation \ref{equ:gprebn}) can impact the optimization of $\gamma, \beta$. We reformulate GpreBN (Equation~\ref{equ:gprebn}) as
\begin{equation}
    {y_i} = \frac{x_i-\mu_c}{\sigma_c}\gamma' + \beta'; \; \gamma' = \frac{\bar{\sigma_c}}{\sigma}\gamma;\; \beta' =  \frac{\bar{\mu_c} - \mu}{\sigma}\gamma + \beta.
    \label{eq7}
\end{equation}
Comparing the standard BN (${y_i} = \frac{x_i - \mu}{\sigma} \gamma + \beta$) and GpreBN (Equation~\ref{eq7}), it is apparent that $\mu, \sigma$ change the initial value in each gradient step. The value of $\mu, \sigma$ should keep $\gamma'$ and $\beta'$ stable in optimization.
Therefore, we expect $\mu, \sigma$ to be a dataset-level value or a special trivial case: $\mu = \bar{\mu}_c, \sigma = \bar{\sigma}_c$. 
In addition, GpreBN allows for a flexible selection of normalization statistics. Inspired by the training process, we estimate a test-time dataset-level running statistics rather than using only current batch statistics. The source statistics are also can be an option since they are dataset-level.

Similar to training phase, in test-time, we using running average to estimate a running mean vector $\mu_r^t$ and running variance vector $\sigma_r^t$ as the testing running statistics. There are two choices for moving average: cumulative moving average (CMA) and exponential moving average (EMA). Here we choose CMA as an example to estimate running statistics as
\begin{equation}
    \mu_r^t = \lambda \mu_c + (1-\lambda) \mu_r^t;\; (\sigma_r^t)^2 = \lambda \sigma_c^2 + (1-\lambda) (\sigma_r^t)^2,
    \label{equ:tr-gprebn}
\end{equation}
where $\lambda = 1/n$, and $n$ is the number of testing batch tracked. 

As proposed in previous work, when the target dataset is too small to get stable statistics~\cite{schneider2020improving} or the target statistics mismatch with the model parameters~\cite{you2021test}, one can mix the current batch statistics with the source running average statistics to stable the estimation or alleviate the mismatch issue. However, the current batch statistics are local, while the source running average statistics are global. We argue that it is better to mix two dataset-level statistics for more stable optimization. Therefore, we mix the source running statistics $\mu_r,\sigma_r$ and testing running statistics $\mu^t_r,\sigma^t_r$ as
\begin{equation}
    \mu = \theta \mu_r^t + (1-\theta) \mu_r;\; \sigma^2 = {\theta (\sigma_r^t)^2 + (1-\theta) \sigma_r^2},
    \label{equ:theta-gprebn}
\end{equation}
where $\theta\in [0,1]$ is a hyper-parameter of the weighting coefficient. We name this statistics option as $\theta$-Mixture.

\section{Experiments}
We conduct our experiments on domain generalization (DG) and robustness, respectively. Our experiments are based on the following Benchmarks: DomainBed \cite{gulrajani2020search} and RobustBench~\cite{croce2020robustbench}. For domain generalization, we evaluate GpreBN on our widely used datasets: PACS~\cite{li2017deeper} (4 domains, 7 classes), OfficeHome~\cite{venkateswara2017deep} (4 domains, 65 classes), VLCS~\cite{fang2013unbiased} (4 domains, 10 classes) and TerraIncognita~\cite{beery2018recognition} (4 domains, 10 classes). For robustness, we evaluate GpreBN on two popular datasets: CIFAR-10-C (10 classes, 15 corruption types, and 5 security levels), CIFAR-100-C~\cite{krizhevsky2009learning} (100 classes, 15 corruption types, and 5 security levels),  ImageNet-C~\cite{deng2009imagenet} (1000 class, 15 corruption types, and 5 security levels). The details of the datasets are provided in Appendix A.

\subsection{Domain Generalization}
In this section, we use the widely used ResNet50~\cite{he2016deep} as the backbone, which is the default setting of DomainBed~\cite{gulrajani2020search} and many other works. We follow the evaluation well acceptable protocol of Domainbed. We compare GpreBN with domain generalization algorithms and other test-time adaptation algorithms. Following You et al.~\cite{you2021test}, which uses mixed statistics in domain generalization, we use our $\theta$-Mixture as the normalization statistics of GpreBN on this task. And we follow Iwasawa et al.~\cite{iwasawa2021test} to select $\theta$ on validation set.

\textbf{DG Baselines} For domain generalization algorithms, following recent works DomainBed~\cite{gulrajani2020search} and SWAD~\cite{cha2021swad}, we choose baselines to include: ERM~\cite{de2018statistical}, GroupDRO~\cite{sagawa2019distributionally}, I-Mixup~\cite{xu2020adversarial,wang2020heterogeneous,yan2020improve}, CORAL~\cite{sun2016deep}, MMD~\cite{li2018domain}, IRM~\cite{arjovsky2019invariant}, ARM~\cite{zhang2020adaptive}, MTL~\cite{blanchard2017domain}, SagNet~\cite{nam2021reducing}, RSC~\cite{huang2020self}, Mixstyle~\cite{zhou2021domain}, and SWAD~\cite{cha2021swad}.

\textbf{Test-time Baselines} Besides the DG baselines, we also compare GpreBN with previous test-time adaptation methods. These baselines include BN-based methods and non-BN-based methods. The BN-based methods are Tent~\cite{wang2020tent} and $\alpha$-BN~\cite{you2021test}. Non-BN-based methods include SHOT~\cite{liang2020we}, PL-C~\cite{lee2013pseudo}, T3A~\cite{iwasawa2021test}. We also compare these baselines on Backbone trained with SWAD~\cite{cha2021swad} (the state-of-the-art DG method). 

\textbf{Hyper-parameters selection protocol} Note that these test-time methods have their own hyper-parameters. Tent has two primary hyper-parameters: $\rho$ for the multiplier of learning rate. $\delta$ for the number of steps updating for each batch, and T3A has one hyper-parameter $M$ for deciding the number of supports to restore. Similarly, GpreBN also has three hyper-parameters: $\rho, \delta$ are the same with tent and $\theta$ for weighting coefficient of weighting statistic. We follow T3A to select hyper-parameters by average accuracy on the validation set. The details of hyper-parameters are provided in Appendix C.

\textbf{Adaptation of Classifier}
Although the proposed GpreBN adapts the feature extractor to the target distribution, the classifier and the decision boundary remain unchanged. Since the classifier never takes the features with GpreBN as the input in the training phase, the features are out-of-domain to some extent. Therefore, adapting the classifier is complementary to GpreBN and should lead to better performance. 
Inspired by T3A, we also adapt the classifier. 
However, different from T3A, we propose a simplified and more practical version by using moving average to estimate the class centroid. Specifically, the $i$-th class centroid, denoted $c_i$, is estimated as
\begin{equation}
    c_i = \frac{N_i}{N_i+m}c_i +  \frac{m}{N_i+m}\bar{z_i};\; p_i = \frac{\exp(z\cdot c_i)}{\sum_{i=1}^C\exp(z\cdot c_i)}
    \label{eq11}
\end{equation}
where $N_i$ is the cumulative number of samples with the pseudo label of the $i$-th class in the testing phase. And $m$ is the number of samples with the pseudo label of the $i$-th class in the current batch. $C$ is the number of classes. $\bar{z}$ is the average of features whose pseudo label is $i$-th class.
${z}$ is the current feature, and $p_i$ is the probability of belonging to $i$-th class.  
The design here lets us only need to record a vector $c_i$ and a samples number $N_i$ for each class, rather than a bank of vectors for each class. In addition, no ranking is needed. We denote the modified adaption of the classifier as rT3A.

\begin{table*}[t]
    \caption{Comparison on four typical domain generalization datasets: PACS, VLCS, OfficeHome, and TerraInc. Both the state-of-the-art DG methods and test-time methods are evaluated on DomainBed~\cite{gulrajani2020search}. The best results are highlighted by bolding. The detailed results are provided in Appendix B.}
    \centering
    \begin{tabular}{lcccc|c}
    \hline
    \textbf{Algorithm}           &\quad PACS \quad&\quad VLCS \quad&\quad OfficeHome \quad&\quad TerraInc  \quad&\quad \textbf{Avg.} \\
    \hline
    ERM       &  85.5   &   77.5   &  66.5    & 46.1 & 68.90  \\
    GroupDRO       &  84.4   &   76.5   &  66.0    & 43.2 & 67.80  \\
    I-Mixup       &  84.6   &   77.4   &  68.1    & 47.9 & 69.38  \\
    CORAL       &  86.2   &   78.8   &  68.7    & 47.7 & 70.05  \\
    IRM       &  83.5   &   78.6   &  64.3    & 47.6 & 68.58  \\
    MMD       &  84.7   &   77.5   &  66.4    & 46.7 & 68.84  \\
    ARM       &  85.1   &   77.6   &  64.8    & 45.5 & 68.38  \\
    MTL       &  84.6   &   77.2   &  66.4    & 45.6 & 68.54  \\
    SagNet       &  86.3   &   77.8   &  68.1    & 48.6 & 69.94  \\
    RSC       &  85.2   &   77.1   &  65.5    & 46.6 & 68.66  \\
    Mixstyle       &  85.2   &   77.9   &  60.4    & 44.0 & 67.28  \\
    SWAD       &  88.1   &   79.1   &  70.6    & 50.0 & 71.34  \\
    \hline
    ERM (reproduced)       &  $84.7$ \scriptsize $\pm 0.5$ & $77.5$ \scriptsize $\pm 1.0$ & $64.9$ \scriptsize $\pm 0.3$ & $46.9$ \scriptsize $\pm 1.8$ & 68.48  \\
    ERM + Tent      &  $\bm{90.2}$ \scriptsize $\pm 0.3$ & $57.8$ \scriptsize $\pm 2.6$ & $61.4$ \scriptsize $\pm 0.2$ & $27.4$ \scriptsize $\pm 3.0$ & 59.18  \\
    ERM + $\alpha$-BN      &  $87.6$ \scriptsize $\pm 0.5$ & $78.3$ \scriptsize $\pm 1.8$ & $62.2$ \scriptsize $\pm 0.2$ & $38.0$ \scriptsize $\pm 4.3$ & 66.54  \\
    ERM + PL-C       &  $82.4$ \scriptsize $\pm 2.1$ & $76.6$ \scriptsize $\pm 1.2$ & $64.4$ \scriptsize $\pm 0.6$ & $44.8$ \scriptsize $\pm 2.3$ & 67.07  \\
    ERM + SHOT       &  $86.3$ \scriptsize $\pm 0.9$ & $73.1$ \scriptsize $\pm 0.7$ & $66.6$ \scriptsize $\pm 0.3$ & $39.9$ \scriptsize $\pm 1.7$ & 66.49  \\
    ERM + T3A       &  $84.6$ \scriptsize $\pm 1.0$ & $79.0$ \scriptsize $\pm 0.4$ & $66.8$ \scriptsize $\pm 0.2$ & $47.0$ \scriptsize $\pm 1.3$ & 69.37  \\
    ERM + GpreBN       &  $88.9$ \scriptsize $\pm 0.6$ & $\bm{80.3}$ \scriptsize $\pm 1.3$ & $66.3$ \scriptsize $\pm 0.6$ & $45.7$ \scriptsize $\pm 2.7$ & $\bm{70.29}$  \\
    ERM + GpreBN + rT3A       &  $86.2$ \scriptsize $\pm 0.4$ & $79.8$ \scriptsize $\pm 0.6$ & $\bm{67.4}$ \scriptsize $\pm 0.3$ & $\bm{47.6}$ \scriptsize $\pm 1.8$ & 70.25  \\
    \hline
    SWAD (reproduced)       &  $88.3$ \scriptsize $\pm 0.2$ & $78.9$ \scriptsize $\pm 0.2$ & $70.6$ \scriptsize $\pm 0.1$ & $50.5$ \scriptsize $\pm 0.3$ & 72.07  \\
    SWAD + Tent      &  \textbf{91.3} \scriptsize $\pm 0.2$ & $56.9$ \scriptsize $\pm 1.0$ & $63.1$ \scriptsize $\pm 0.1$ & $31.0$ \scriptsize $\pm 1.5$ & 60.57  \\
    SWAD + $\alpha$-BN      &  $89.0$ \scriptsize $\pm 0.6$ & $78.2$ \scriptsize $\pm 3.1$ & $66.7$ \scriptsize $\pm 3.0$ & $45.4$ \scriptsize $\pm 1.3$ & 69.81  \\
    SWAD + PL-C       &  $88.1$ \scriptsize $\pm 0.2$ & $76.9$ \scriptsize $\pm 1.8$ & $70.1$ \scriptsize $\pm 0.6$ & $51.4$ \scriptsize $\pm 0.3$ & 71.65  \\
    SWAD + SHOT       &  $89.5$ \scriptsize $\pm 0.3$ & $76.2$ \scriptsize $\pm 0.0$ & $71.5$ \scriptsize $\pm 0.3$ & $41.7$ \scriptsize $\pm 0.5$ & 69.76  \\
    SWAD + T3A       &  $88.6$ \scriptsize $\pm 0.2$ & $81.0$ \scriptsize $\pm 0.2$ & $71.9$ \scriptsize $\pm 0.3$ & $50.5$ \scriptsize $\pm 0.5$ & 73.00  \\
    SWAD + GpreBN       &  $90.5$ \scriptsize $\pm 0.2$ & $\bm{82.1}$ \scriptsize $\pm 0.4$ & $71.1$ \scriptsize $\pm 0.2$ & $49.6$ \scriptsize $\pm 0.9$ & $73.31$  \\
    SWAD + GpreBN + rT3A       &  $89.4$ \scriptsize $\pm 0.2$ & $81.2$ \scriptsize $\pm 0.3$ & $\bm{72.2}$ \scriptsize $\pm 0.2$ & $\bm{51.8}$ \scriptsize $\pm 0.5$ & $\bm{73.62}$  \\
    \hline
    \end{tabular}
    \label{tab:dg_results}
    \vspace{-1em}
\end{table*}

\textbf{Results} The results are presented in Tabel~\ref{tab:dg_results}, from which we can see that the proposed GpreBN with ERM can already outperform all of the non-test time methods except SWAD. In addition, GpreBN outperforms all of the other test-time methods. Though ERM + GpreBN does not outperform SWAD, GpreBN consistently improves the accuracy when applied to SWAD. By applying GpreBN to SWAD, we achieve state-of-the-art performance on domain generalization. Specifically, T3A works better on OfficeHome and TerraInc, and GpreBN works better on PACS and VLCS. The reason could be that although the feature extractor is adapted, the classifier may still mismatch on OfficeHome and TerraInc. Considering that, we apply rT3A to our method and find that in this case, GpreBN+rT3A can outperform T3A on OfficeHome and TerraInc. We also observe that after adaptation using Tent, the performance drops a lot (ERM: from 68.9\% to 59.18\%, SWAD: from 72.07\% to 60.57\%). We suppose the reason is that the model is fine-tuned on the source datasets using Imagenet pre-trained initialization with BN layer frozen. The statistics of frozen BN are the estimated statistics on Imagenet. 
In this case, the features are normalized by constant vectors rather than their own statistics. This leads to out-of-domain when Tent switches the constant vectors to batch-specific statistics. Please see the detailed analysis in Section~\ref{sec:analysis}.

\subsection{Robustness}
In this section, we verify the effectiveness of GpreBN on the robustness task. The experiments in this section are based on Robustbench~\cite{croce2020robustbench}, which is a standardized adversarial robustness benchmark. We use 40-2 WideResNet (Wide Residual Network)~\cite{zagoruyko2016wide} trained with AM~\cite{hendrycks2019augmix} as the backbone for CIFAR-10-C and  CIFAR-100-C. We use ResNet50 trained with AM~\cite{hendrycks2019augmix} as the backbone for IamgeNet-C, whose checkpoints are provided by Robustbench. Similar to the experiments in domain generalization, we compare GpreBN to robustness algorithms and other test-time adaptation methods. Since using $\theta$-Mixture for GpreBN introduces a new hyperparameter $\theta$, the selection of $\theta$ will result in unfair comparison in robustness. Considering Tent~\cite{wang2020tent} uses the current testing statistics (target statistics) for robustness, we use testing running statistics (TRS, Equation~\ref{equ:tr-gprebn}) for GpreBN instead, which are our proposed testing statistics.

\textbf{Robustness baselines.} Since AM also conducts their experiments based on Robustbench, we mainly compare our methods with the results reported in~\cite{hendrycks2019augmix}. Follow hendrycks et al.~\cite{hendrycks2019augmix}, We compare our methods with the following augmentation-based methods on CIFAR-10-C and CIFAR-100-C: Cutout~\cite{devries2017improved}, Mixup~\cite{zhang2017mixup}, CM~\cite{yun2019cutmix}, AutoAugment (AA)~\cite{cubuk2018autoaugment}, Adv Training~\cite{madry2017towards},  and AM~\cite{hendrycks2019augmix}. For ImageNet-C, we compare our methods with the following baselines: AA~\cite{cubuk2018autoaugment}, Random AA~\cite{hendrycks2019augmix}, MaxBlur pool~\cite{zhang2019making} and Stylized ImageNet (SIN)~\cite{geirhos2018imagenet}.

\textbf{Test-time baselines.} We also compare our methods with other test-time methods. The baselines include T3A~\cite{iwasawa2021test}, Tent~\cite{wang2020tent}, BN~\cite{schneider2020improving} (using statics of current batch without optimization).

\textbf{Results} We follow Tent~\cite{wang2020tent} to evaluate these test-time methods. For a fair comparison, we use the same hyper-parameters with Tent~\cite{wang2020tent} for GpreBN, which are the same for all three datasets. The results of CIFAR-10-C and CIFAR-100-C are presented in Table~\ref{tab:rb_cifar_results}. Our method consistently works better than all of the baselines. The detailed results on CIFAR-10-C and CIFAR-100-C are demonstrated in Table~\ref{tab:rb_cifar_detailed}, from which we can see that our model constantly achieves better performance for different corruption types. 
We present the results of ImageNet-C in Table~\ref{tab:rb_im_results} \footnote{The mean corruption error (mCE) is different from CIFAR-10-C or CIFAR-100-C. We follow Hendrycks et al.~\cite{hendrycks2019augmix} to calculate it.}. Our method outperforms the baselines consistently for different corruption types.
Note that T3A is not as effective as in domain generalization and even hurts the performance on ImageNet-C. We deduce that the reason may be that, in robustness, the shift of testing and training data is caused by small perturbations, which destroy the low-level feature but maintain the high-level feature. Considering that the classifier is in charge of high-level semantics, the major mismatch lies in the feature extractor.

\begin{table*}[t]
\small
    \caption{Comparison on CIFAR-10-C and CIFAR-100-C dataset. Both the robustness methods and test-time methods are evaluated on Robustbench~\cite{croce2020robustbench}. The best results are highlighted by bolding. mCE is reported (lower is better).}
    \centering
    \begin{tabular}{lccccc|cccccc}
    \hline
    \textbf{Datasets}   &Cutout&Mixup&CM&AA&Adv&AM&+Tent&+T3A&+BN& +GpreBN\\
    \hline
    CIFAR-10-C       &  26.8   &   22.3   &  27.1    & 23.9 & 26.2 &11.2 &9.2&10.7 & 10.8 &\textbf{8.9}\\
    \hline
    CIFAR-100-C       &  53.5   &   50.4   &  52.9    & 49.6 & 55.1 & 35.9 & 31.2 & 35.6 &34.1&\textbf{30.6} \\
    \hline
    \end{tabular}
    \label{tab:rb_cifar_results}
\end{table*}

\begin{table*}[t]
    \caption{The detailed comparision on CIFAR-10-C and CIFAR-100-C dataset, including 15 corruption types and 5 security levels. The best results are highlighted by bolding. mCE is reported (lower is better).}
    \centering
    \begin{tabular}{c|ccccc|ccccc}
    \hline  
    Type&AM&+Tent&+T3A&+BN&+GpreBN&AM&+Tent&+T3A&+BN&+GpreBN\\ 
    \hline
    Gauss.&19.1&11.7&17.3&14.1& \textbf{11.3}&52.1&35.5&51.1&39.1& \textbf{35.0}\\ 
    Shot&14.0&9.9&13.0&11.9& \textbf{9.6}&44.3&33.2&43.9&36.6& \textbf{32.6}\\ 
    Impulse&13.3&11.7&13.0&13.9& \textbf{11.5}&40.3&33.3&40.4&37.1& \textbf{32.8}\\ 
    Defocus&6.3&6.3&6.3&7.2& \textbf{6.1}&26.6&26.7&27.0&28.7& \textbf{26.1}\\ 
    Glass&17.1&14.7&16.4&17.6& \textbf{14.3}&45.9&37.4&44.2&41.0& \textbf{36.7}\\ 
    Motion&7.9&7.6&7.8&8.7& \textbf{7.4}&29.5&28.4&29.8&30.6& \textbf{28.0}\\ 
    Zoom&7.0&6.7&6.8&7.9& \textbf{6.6}&28.2&27.6&28.4&30.3& \textbf{27.0}\\ 
    Snow&10.4&9.1&10.3&10.8& \textbf{8.9}&34.2&31.6&34.4&34.7& \textbf{31.2}\\ 
    Frost&10.6&8.7&10.4&10.6& \textbf{8.5}&36.3&31.3&36.3&34.2& \textbf{30.8}\\ 
    Fog&8.5&7.7&8.0&9.1& \textbf{7.5}&33.4&30.5&33.1&34.1& \textbf{29.8}\\ 
    Bright&5.9&6.0&5.9&6.8& \textbf{5.8}&26.5&26.0&26.9&28.1& \textbf{25.5}\\ 
    Contr.&9.7&7.6&9.2&9.0& \textbf{7.3}&34.7&28.6&33.9&31.2& \textbf{28.0}\\ 
    Elastic&9.2&9.7&9.1&10.9& \textbf{9.4}&31.9&31.6&32.1&34.2& \textbf{31.0}\\ 
    Pixel&16.8&8.2&15.0&10.1& \textbf{7.9}&36.5&28.9&35.8&31.8& \textbf{28.2}\\ 
    JPEG&11.9&12.1&11.9&14.0& \textbf{11.8}&37.9&36.8&38.1&40.0& \textbf{36.2}\\ 
    \hline
    \textbf{mCE.}&11.2&9.2&10.7&10.8& \textbf{8.9}&35.9&31.2&35.6&34.1& \textbf{30.6}\\  
    \hline
    \end{tabular}
    \label{tab:rb_cifar_detailed}
    \vspace{-1em}
\end{table*}

\begin{table*}[t]
    \caption{Comparison on ImageNet-C dataset. Both the robustness methods and test-time methods are evaluated on Robustbench~\cite{croce2020robustbench}.The best results are highlight by bolding.  The metric is mean corruption errer (lower is better). The best results are highlight by bold.}
    \centering
    \begin{tabular}{lcccc|c}
    \hline
    \textbf{Algorithm}           &\quad Noise \quad&\quad Blur \quad&\quad Weather \quad&\quad Digital\quad&\quad \textbf{mCE.} \\
    \hline
    AutoAugment       &  69.7 & 80.2 & 68.5 & 71.5 & 72.7  \\
    Random AA       &  71.0 & 82.2 & 72.8 & 77.0 & 66.1  \\
    MaxBlur pool       &  74.3 & 78.8 & 67.0 & 74.0 & 73.4  \\
    SIN       &  69.7 & 79.8 & 70.8 & 72.5 & 73.3\\
    AM       &  66.0 & 70.5 & 68.0 & 68.8 & 68.4  \\
    \hline
    AM(reproduced)       &  65.2 & 68.0 & 65.7 & 64.1 & 65.8  \\
    +Tent       &  49.7 & 50.5 & 46.4 & 48.7 & 48.7  \\
    +T3A       &  65.4 & 69.1 & 66.3 & 64.5 & 66.3  \\
    +BN       &  56.0 & 56.4 & 50.0 & 52.7 & 68.9  \\
    +GpreBN       &  \textbf{49.2} & \textbf{50.0} & \textbf{46.0} & \textbf{48.3} & \textbf{48.3}  \\
    \hline
    \end{tabular}
    \label{tab:rb_im_results}
\end{table*}

\subsection{Ablation Studies}
\label{Sec:ablation}
In order to verify the effectiveness of the proposed GpreBN, we conduct the following ablation studies on both domain generalization and robustness tasks. Our ablation study is conducted on PACS for domain generalization. Besides, we conduct an ablation study on CIFAR-10-C for robustness.

\textbf{Ablation study on domain generalization} We locate the effectiveness of domain generalization of GpreBN in two factors: gradient preserved normalization and testing running statistics, which correspond to our two insights. 

\begin{table*}[t]
    \caption{Ablation study on gradient preserving (GradPre.). The methods shares hyperparameters. All the methods are based on the same ERM models trained on the PACS dataset. We use a checkmark in Opt to denote minimizing entropy.}
    \centering
    \begin{tabular}{cc|c|cccc|c}
    \hline
    BN type   &   Statistics & GradPre.& Art. & Cartoon & Photo & Sketch& Avg. \\
    \hline
    BN& $\mu_r$ & &  86.4&76.2&95.7&80.3&84.6\\
    GpreBN   & $\mu_r$  &\CheckmarkBold &  85.5&79.6&95.6&82.7&85.8  \\
    BN(de) & $\mu_c$  & & 88.1&84.9&97.3&80.3&87.6  \\
    BN  & $\mu_c$ & \CheckmarkBold &  91.3&87.8&98.1&83.3&90.2  \\
    \hline
    \end{tabular}
    \label{Ab_dg:factor1}
    \vspace{-1em}
\end{table*}

For the first factor, gradient preserved normalization, we conduct the following experiment to verify the effectiveness. 
The normalization statistics $\mu,\sigma$ can take either $(i)$ the source statistics ($\mu = \mu_r, \sigma = \sigma_r$) or $(ii)$ target statistics ($\mu = \mu_c, \sigma = \sigma_c$).  Since the statistics used to normalize features may impact the performance, both cases should be considered.  For the first case, we compare the model that uses GpreBN (Equation \ref{equ:gprebn}) to the traditional BN (${y_i} = (x_i - \mu)/\sigma \cdot \gamma + \beta$) in the test-time scenario. 
For the second case, we stop the gradient of $\mu,\sigma$ for BN (denoted as BN(de)) when minimizing entropy to verify our assumption.
From Table~\ref{Ab_dg:factor1}, we know that gradient preserving is effective in both cases. With gradient preserved normalization, GpreBN ($\mu_r$) significantly improves the accuracy compared to BN ($\mu_r$). The performance of BN(de) ($\mu_c$) significantly drops compared to BN ($\mu_c$) due to lacking gradient preserved normalization.


For the second factor, testing running statistics ($\mu = \mu_r^t, \sigma = \sigma_r^t$), we compare it to using target statistics of the current batch ($\mu = \mu_c, \sigma = \sigma_c$). The comparison is under the cases of with or without entropy minimizing. From Table~\ref{Ab_dg:factor2}, we can find that BN ($\mu_r^t$) outperforms BN ($\mu_t$) without minimizing entropy. GpreBN ($\mu_r^t$) outperforms GpreBN ($\mu_c$) with entropy minimizing. To conclude, testing running statistics are better than target statistics of the current batch in both cases with or without minimizing entropy. We also study deriving running statistics methods: using cumulative moving average (CMA) or exponential running average (EMA). We see that using EMA is comparable or slightly better than CMA. 

\begin{table*}[t]
    \caption{Ablation study on testing running statistics (TRS). The methods shares the same hyperparameters. All the methods are based on the same ERM models trained on the PACS dataset. Opt represent the method minimize entropy (checkmark) or not. }
    \centering
    \begin{tabular}{cc|cc|cccc|c}
    \hline
    BN type   &   Statistics &    Opt & TRS & Art. & Cartoon & Photo & Sketch& Avg. \\
    \hline
    BN  & $\mu_c$&    &  & 85.9&82.7&96.9&76.8&85.6  \\
    BN  & $\mu_r^t$ &  &  \CheckmarkBold &  86.4&83.4&97.2&77.0&86.0  \\
    GpreBN  & $\mu_c$ &  \CheckmarkBold &   &  91.3&87.8&98.1&83.3&90.2  \\
    GpreBN & $\mu_r^t$ &  \CheckmarkBold &   \CheckmarkBold &  91.9&88.9&98.3&82.7&90.5  \\
    GpreBN(EMA) & $\mu_r^t$ &  \CheckmarkBold  &  \CheckmarkBold & 92.0&88.6&98.3&83.5&90.6  \\
    \hline
    \end{tabular}
    \label{Ab_dg:factor2}
\end{table*}

\textbf{Ablation study on robustness} The two key points for our method on robustness are the same with domain generalization: $(i)$ gradient preserved normalization, $(ii)$ test running statistics. 

For verification of the first point, we consider the case of using source statistics or using target statistics. In addition, we also verify the effectiveness by stopping the gradient of target statistics. As Table \ref{ab_rb_table} shows, when using the source statistics GpreBN ($\mu_r$) outperforms BN ($\mu_r$). On the contrary, we stop the gradient of $\mu,\sigma$ for BN (denoted as BN(de)) to verify our assumption of the effectiveness of the cross-instance backpropagation. From Table \ref{ab_rb_table}, the performance of BN(de) significant drops comparing with BN ($\mu_c$).

The effectiveness of the running statistics has been verified in the main results of robustness. However, the influence baseline robustness method has not been verified. Therefore, we also apply our method to the standard model. As shown in Table \ref{ab_rb_table}, using running statistics (TRS), GpreBN ($\mu_r^t$) outperforms BN ($\mu_c$, with out TRS) for both two models. We also compare the cumulative moving average (GpreBN) and exponential running average (GpreBN(EMA)). As shown in Table \ref{ab_rb_table}, GpreBN(EMA) achieves a similar performance as GpreBN.


\begin{table*}[t]
    \caption{Ablation study on gradient preserving (GradPre.) and testing running statistics (TRS) for robustness. The methods share the same hyperparameters. All the methods are based on the same ERM models trained on the PACS dataset. mCE is reported (lower is better).}
    \centering
    \begin{tabular}{cc|cc|cc}
    \hline
    BN type   &   Statistics & GradPre.& TRS & Standard & AM\\
    \hline
    BN  & $\mu_r$  &  & &  42.1&17.7  \\
    GpreBN & $\mu_r$  & \CheckmarkBold & & 27.3&10.2 \\
    BN(de) & $\mu_c$  &  &  &  13.8&10.3  \\
    BN & $\mu_c$  & \CheckmarkBold &  &   13.2 & 9.2 \\
    GpreBN & $\mu_r^t$   & \CheckmarkBold & \CheckmarkBold & 13.1&8.9 \\
    GpreBN(EMA) & $\mu_r^t$  & \CheckmarkBold & \CheckmarkBold & 13.0&8.9 \\
    \hline
    \end{tabular}
    \label{ab_rb_table}
\end{table*}


\subsection{Analysis}
\label{sec:analysis}
From the results above, it can be concluded that GpreBN is the only method effective for both domain generalization and robustness. Tent fails on domain generalization (VLCS, OfficeHome, and TerraInc), and T3A is not so effective on robustness as on domain generalization. In this section, we try to analyze the reasons for these two failure cases. 

\textbf{The reason why Tent fails on domain generalization.} To illustrate the problem, we take VLCS as an example. From Table~\ref{tab:studyType}, we know that target statistics fail on VLCS. While Tent uses target statistics, which is a baseline with poor performance. Therefore, the major cause of why Tent failed is that target statistics failed in VLCS. Since Tent works well on robustness (using a pre-trained model) but fails on domain generalization (using a fine-tuned model with BN layer fixed, which is pre-trained on ImageNet). We suppose that the target statistics failed is due to the fine-tuning on the source domains. In order to verify the statement above, we conduct the following experiments: $(i)$ For domain adaptation, we fix the extractor but only fine-tune the classifier on source domains (using pre-trained models). $(ii)$ For robustness, we fine-tune the model on 14 corruption types but test on the left one (using the fine-tuned model with BN layer fixed). 
The results are presented in Table~\ref{tab:studyType}, from which we can see that using source statistics is better in the fine-tuned setting, and using target statistics is better in the pre-trained setting on both domain generalization and robustness.

\begin{table}[t]
    \caption{Performance under different settings. We report average accuracy(Avg. $\uparrow$) for VLCS and mean corruption error(mCE $\downarrow$) for CIFAR-10-C. Typically, domain generalization uses the fine-tuned model, and robustness uses the pre-trained model.}
    \centering
    \begin{tabular}{cc|cc|cc}
    \hline
    Setting   & \quad  Statistics \quad&\quad  VLCS\quad & \quad+Tent \quad& \quad CIFAR-10-C \quad & \quad +Tent \\
    \hline
    \multirow{2}{*}{fine-tuned}  & $\mu_r$& \quad \textbf{77.5}  & \textbf{77.2} & \textbf{4.3} & 4.6 \\
    & $\mu_c$ &  \quad56.4 &  57.8 &4.8 & 3.6 \\
    \hline
    \multirow{2}{*}{pre-trained}  & $\mu_r$& \quad 70.0  &73.0   &43.5 & 42.1\\
    & $\mu_c$ &\quad \textbf{70.1}  & \textbf{74.2} & \textbf{14.4} & \textbf{13.2} \\
    \hline
    \end{tabular}
    \label{tab:studyType}
\end{table}

\textbf{The reason why T3A not so effective for robustness.} T3A only adjusts the classifier but fixes the feature extractor. However, if the reason for out of domain is mainly on the feature extractor, the improvement of adjusting the classifier is quite limited. Intuitively, the distribution shift of robustness is caused by the small perturbation of images, which destroys the low-level features of data but maintains the high-level semantics. Therefore, we suppose that the out of distribution of robustness is mainly in the early layer of networks. 
We conduct the following experiment to verify our idea. Instead of using target statistics for all the BN layers (named \emph{all target}), we only use target statistics at the 50\% earlier BN layers of the model, and the other layers remain unchanged (named \emph{50\% target}). 
Compared to \emph{all target} (error: 14.4), \emph{50\% target} (error: 14.6) only causes slightly drop. This verifies our idea that out of domain is mainly on the feature extractor, and T3A is not effective for robustness.

Based on the analysis above, compared to Tent, GpreBN is flexible to make a better choice of normalization statistics and maintain effectiveness (preserving the gradient backpropagation). GpreBN focuses on the adaptation of feature classifiers and is complementary with T3A. We also provide a more practical rT3A. In summary, GpreBN is a general technique for handling distribution shift and can be applied to both domain generalization and robustness.

\section{Conclusion}
We investigate batch normalization in test time and propose two crucial insights for test-time BN in handling distribution shift: $(i)$ cross instance gradient backpropagation of BN layer and $(ii)$ dataset-level normalization statistics. We present a simple-to-implement GpreBN layer for test-time adaptation based on these insights. Our experiments show that our proposed GpreBN achieves state-of-the-art performance on both domain generalization and robustness. Besides, our ablation study demonstrates the effectiveness of each of our proposed techniques. We also provide a classifier adaptation technique rT3A. Currently, we can only adapt the networks with BN layers. In the future, we would like to extend our idea to other types of normalization, e.g., layer norm, and apply it to other network structures.

\bibliographystyle{plain}
\bibliography{aimc2022}
\end{document}